\documentclass[runningheads]{llncs}

\usepackage{graphicx}
\usepackage{amsmath}
\usepackage{amssymb}
\usepackage{amsthm}
\usepackage{multirow}
\usepackage{multicol}
\usepackage{threeparttable}
\theoremstyle{definition}

\usepackage[ruled,linesnumbered]{algorithm2e}
\usepackage{xcolor}

\begin{document}

\copyright 20xx IEEE. Personal use of this material is permitted. Permission
from IEEE must be obtained for all other uses, in any current or future
media, including reprinting/republishing this material for advertising or
promotional purposes, creating new collective works, for resale or
redistribution to servers or lists, or reuse of any copyrighted
component of this work in other works.

\newpage

\title{SAS: A Simple, Accurate and Scalable Node Classification Algorithm}
%
%
\author{Ziyuan Wang\thanks{Contributed equally to this work.} \and
Feiming Yang\inst{\star} \and
Rui Fan}
\authorrunning{Z. Wang, F. Yang, and R. Fan}
%
\institute{School of Information Science and Technology, ShanghaiTech University
\email{\{wangzy11,yangfm,fanrui\}@shanghaitech.edu.cn}}


%
\maketitle              
\begin{abstract}
Graph neural networks have achieved state-of-the-art accuracy for graph node classification. However, GNNs are difficult to scale to large graphs, for example frequently encountering out-of-memory errors on even moderate size graphs. Recent works have sought to address this problem using a two-stage approach, which first aggregates data along graph edges, then trains a classifier without using additional graph information. These methods can run on much larger graphs and are orders of magnitude faster than GNNs, but achieve lower classification accuracy. We propose a novel two-stage algorithm based on a simple but effective observation: we should first train a classifier then aggregate, rather than the other way around. We show our algorithm is faster and can handle larger graphs than existing two-stage algorithms, while achieving comparable or higher accuracy than popular GNNs. We also present a theoretical basis to explain our algorithm's improved accuracy, by giving a synthetic nonlinear dataset in which performing aggregation before classification actually decreases accuracy compared to doing classification alone, while our classify then aggregate approach substantially improves accuracy compared to classification alone.

\end{abstract}

\section{Introduction}
One of the most important graph-based learning problems is node classification, which tries to determine the labels of nodes based on node features, the graph topology, and known labels on a subset of the nodes.  Among many methods which have been proposed for this task \cite{roweis2000nonlinear,zhou2004learning,zhu2005semi,perozzi2014deepwalk,grover2016node2vec,kipf2016semi}, graph neural networks (GNNs) have emerged as the most accurate.  However, most GNN algorithms are limited in scalability \cite{kipf2016semi,velivckovic2017graph,xu2018representation,klicpera2018predict,chen2020simple}: they can only process moderate size graphs, \emph{e.g.} with less than one million edges, and run out of memory on larger graphs.  While sampling techniques \cite{hamilton2017inductive,chiang2019cluster,zeng2019graphsaint} have been proposed to address this problem, these algorithms are slower, still have high constant factors in their memory complexity, and are difficult to train.  

A recent approach to improve the computational performance of node classification is to simplify the GNN architecture.  In particular, whereas GNNs are trained end-to-end using forward steps and backpropagation through multiple graph layers, new \emph{two-stage} approaches \cite{wu2019simplifying,nt2019revisiting} perform two separate steps, \emph{aggregation} of node features or latent representations along graph edges, and then learning \emph{transformations} on the aggregated features without using any additional graph information.  This greatly improves speed and memory consumption, and two stage algorithms can scale to orders of magnitude larger graphs.  However, these algorithms also have lower accuracy compared to GNNs, especially on the large graphs that they were designed to deal with.  

In this work we present a novel two-stage algorithm which achieves the same or better scalability as existing two-stage algorithms, while also achieving competitive or higher accuracy than state-of-the-art GNNs.  Our algorithm is based on a simple but important observation: instead of performing aggregation followed by feature transformation as in current algorithms, we can improve accuracy and performance by performing transformation first and then aggregating.  We demonstrate the effectiveness of this approach on a number of real world datasets, including large graphs with over one hundred millions edges.  We also provide a theoretical basis to explain why the order of transformation and aggregation affects accuracy.  In particular, we construct a nonlinear dataset on which we evaluate permutations of multiple aggregation and transformation steps, and show that aggregating before transforming actually decreases accuracy compared to performing only transformations without aggregation, whereas aggregating after feature transformation significantly improves accuracy compared to transforming alone.

In the remainder of the paper, we first review related works on node classification.  We then provide some algorithmic background for GNNs, two-stage algorithms and our method.  We present our algorithm in Section \ref{sec-alg} and discusses some of its advantages.  In Section \ref{sec-exp}, we benchmark our algorithm on a number of datasets against other methods, and also present synthetic datasets which motivate our design and explain its performance.  Finally, we conclude in Section \ref{sec-concl}.

\section{Related Works}

A number of algorithms have been proposed to classify nodes based on their features and graph information.  Earlier works often made use of only the features or graph information without combining them, whereas later works based on graph neural networks combined both types of information and achieved higher accuracy.

\subsubsection{Traditional Algorithms}
Label propagation algorithms \cite{zhou2004learning,zhu2005semi} spread one-hot labels from the labeled training nodes to the unlabeled test nodes, but do not perform any feature transformations on nodes.  Node embedding algorithms work by finding low-dimensional vector representations of nodes and using these for classification. For example, LLE \cite{roweis2000nonlinear} performs unsupervised dimensionality reduction on node features without requiring training nodes, and also uses 1-hop graph information. DeepWalk \cite{perozzi2014deepwalk} and Node2vec \cite{grover2016node2vec} are random walk methods which randomly create paths from certain starting nodes and explore the graph structure both locally and globally.  However, these paths are often unstable, especially on large and dense graphs.  These algorithms generally achieve lower accuracy compared to recent graph neural network methods.  
\subsubsection{Graph Neural Networks} Graph convolution networks (GCN) \cite{kipf2016semi} generalize classical convolutional neural networks (CNN) \cite{lecun1995convolutional} to graph structures using aggregation operators to combine latent representations from neighbors, and currently achieve the highest classification accuracy on a number of tasks.  There are also a number of variants of GCNs, for example Graph Attention Networks (GAT) \cite{velivckovic2017graph} which use an attention mechanism during aggregation.  

In traditional neural networks such as CNNs \cite{lecun1995convolutional}, increasing the depth of the network also increases its expressiveness and accuracy.  Surprisingly, the situation for GNNs is different, and shallow GNNs which only work on a few layers of the graph often perform better.  As the network depth increases, nodes tend to \emph{over-smooth}, \emph{i.e.} converge to the same representation, which leads to lower accuracy.  A series of works have tried to address this problem.  JKNet \cite{xu2018representation} adds residual connections from previous layers and learns the weights of these connections. APPNP \cite{klicpera2018predict} uses residual connections to helps nodes converge to their approximate personalized PageRank values.  GCNII \cite{chen2020simple} improves on the convergence properties of APPNP and allows the use of very deep GNNs.  Nevertheless, none of the previous methods demonstrate significantly higher classification accuracy compared to shallow GNNs.  At the same time, they incur substantial performance penalties in terms of training time and memory consumption, which restricts their use to only small scale graphs.  This led to a line of work studying simpler network architectures.  

SGC \cite{wu2019simplifying} proposed that large depth is not a necessary feature for GNNs, and additionally using only linear feature transformations is sufficient.  They developed a simplified GCN model consisting of multi-hop aggregation of node features followed by a simple linear classifier.  Training is done only on the classifier, and the aggregation step is independent of node features.  SIGN \cite{rossi2020sign} uses a similar approach, but concatenates neighbor information from multiple hops.  GfNN \cite{nt2019revisiting} improves SGC by using a deeper classifier after feature aggregation.  These methods are significantly faster than traditional GNNs and consume much less memory because their training can be easily mini-batched.  However, they also have substantially lower accuracy than GNNs, especially on large graphs.  Our work aims to be both highly scalable and accurate. The Correct and Smooth (C\&S) algorithm \cite{huang2020combining} first trains a classifier using a spectral embedding of the graph, corrects the base prediction using aggregation, then does additional aggregation of the one-hot labels of the training nodes.  While this classify-then-aggregate approach is similar to ours, C\&S has limited performance due to its large number of aggregation steps and the need to compute a spectral embedding, and is also limited to transductive classification tasks.  Our algorithm achieves higher performance and can perform transductive and inductive classification.

Several recent works have aimed to deploy GNNs on large graphs \cite{hamilton2017inductive,chen2018fastgcn,zeng2019graphsaint,chiang2019cluster}.  These algorithms use different subgraph sampling techniques to reduce the portion of the input graph which needs to be stored in memory during training.  GraphSAGE \cite{hamilton2017inductive} randomly samples a fixed number of neighbors during each node's propagation step, but the total number of sampled nodes still grows exponentially with the number of layers. \cite{fey2019fast} improves this method using importance sampling. \cite{chiang2019cluster} and \cite{zeng2019graphsaint} use clustering algorithms to preprocess the graph and reduce overlap in the sampled subgraphs, but this step substantially increases the execution time.  These algorithms are significantly slower than our algorithm and other two-stage algorithms described above.

\section{Preliminaries}
We first introduce some background and notation used in the rest of the paper.  Let $G = (V,E)$ be a graph with node set $V$ and an undirected edge set $E$, and let $N = |V|$ be the number of nodes in $G$ and $M = |E|$ be the number of edges.  For node classification problems, each node $i$ has a $D$ dimensional feature vector $\mathbf{x}_i\in \mathbb{R} ^D$, and a $C$ dimensional one-hot label $\mathbf{y}_i\in \{0,1\}^C$, indicating that $i$ belongs to one of $C$ classes.  The input to the classification algorithm consists of a set of features and labels $(\mathbf{x}_i, \mathbf{y}_i)$ for $i \in T$, where $T \subseteq V$ is the set of \emph{training} nodes. By stacking the features and labels of all the nodes, we form a feature matrix $\mathbf{X}\in\mathbb{R}^{N \times D}$ and label matrix $\mathbf{Y}^{N \times C}$.

We consider two types of classification tasks, \emph{transductive} and \emph{inductive}.  In the former, our goal is to classify the unlabeled nodes in a graph, 
\emph{i.e.} nodes in $V \backslash T$.  In the latter task, we train our classification algorithm on the training nodes in $G$, and want to classify unlabeled nodes in $G$ and also nodes in other graphs $G'$ which do not contain any additional training nodes.  

\subsubsection{Graph Convolutional Networks and Variants}

Graph Convolution Networks \cite{kipf2016semi} (GCNs) consist of a $K$-layer neural network combining feature transformation and graph convolution operations. Learning a node representation $\mathbf{H}^{(K)}$ is based on the following recursive definition:
\begin{equation}
    \mathbf{H}^{(0)}=\mathbf{X}, \qquad \mathbf{H}^{(l)}=\sigma(\Tilde{\mathbf{S}}\mathbf{H}^{(l-1)}\mathbf{\Theta}^{(l)}),l\in\{1,\dots,K\}
\end{equation}
Here,  $\mathbf{H}^{(l)}\in \mathbb{R}^{N\times h_l}$ is the hidden representation of nodes at layer $l$, $\mathbf{\Theta}^{(l)}\in \mathbb{R}^{h_l\times h_{l+1}}$ is the learned weight matrix for linearly transforming  $\mathbf{H}^{(l)}$, and $h_0=D,h_K=C$. GCN uses the normalized self-looped adjacency matrix
\begin{equation}
    \label{norm_adj}
    \Tilde{\mathbf{S}}=(\mathbf{D}+\mathbf{I})^{-\frac12}(\mathbf{A}+\mathbf{I})(\mathbf{D}+\mathbf{I})^{-\frac12}
\end{equation}
as the graph convolution operator, where $\mathbf{A}\in\mathbb{R}^{N\times N}$ is the adjacency matrix of $G$, and $\mathbf{D}={\rm diag}(d_1,d_2,\dots,d_N)$ is a matrix with the degrees of the nodes on its diagonal and zeros elsewhere.  $\sigma$ is a nonlinear activation function; in particular, we use softmax in the last layer and ReLU in the other layers.  

A number of variant GNNs exist which are based on a similar structure as GCNs.  These algorithms can be decomposed into two basic operations, which we call \emph{transformation} and \emph{aggregation}.  The first operation transforms the latent representations of the nodes, and the second combines the representations of the neighbors of a node (including the node itself).  For example, in GCN, the transformation step is implemented by the $\sigma$ function and multiplication by $\mathbf{\Theta}$, while aggregation is implemented by multiplication by $\Tilde{\mathbf{S}}$.  By varying the specific functions used for transformation and aggregation and also the order in which these functions are combined, a number of GNNs with different accuracy and performance properties can be created.

\subsubsection{Simplified GNNs}
Standard GNNs combine multiple steps of transformation and aggregation into a single function, then train this function end-to-end based on training labels.  This requires maintaining a large amount of data in memory in order to perform operations such as backpropagation, which in turn limits the size of the graph which can be processed unless complex procedures such as graph sampling or partitioning are applied.  Simple Graph Convolution (SGC) \cite{wu2019simplifying} proposes removing nonlinear activations except the final softmax and combining all the transformations, and can be expressed as the following:
\begin{equation}
 \mathbf{H}^{(0)} = \Tilde{\mathbf{S}}^K  \mathbf{X}, \qquad \mathbf{H}^{(1)}={\rm softmax}( \mathbf{H}^{(0)} \mathbf{\Theta})
\end{equation}
Thus, SGC is divided into two stages, the first consisting of $K$ rounds of aggregation using the initial feature vectors of the nodes, and the second to train a multi-class logistic regression model based on the aggregated features.  The parameter-free propagations $\Tilde{\mathbf{S}}^K \mathbf{X}$ can be computed in a preprocessing step, and the only training is on the parameters $\mathbf{\Theta}$ during the transformation step.  This allows mini-batch training without graph sampling or partitioning and allows SGC to run on much larger graphs.  Despite these advantages, we found in our experiments that SGC suffers from decreased accuracy, especially on large graphs.  As we discuss in Section \ref{sec:sythetic}, we believe two key reasons for this are that SGC performs aggregation before transformation, and also SGC's use of linear instead of nonlinear transformations.

\section{SAS Algorithm}
\label{sec-alg}

In this section we propose SAS, a simple, accurate and scalable node classification algorithm.  SAS aims to combine both features and graph information from the input to obtain higher accuracy than using either type of data alone.  It is based on the assumption that features are not completely informative.  That is, it assumes that even an optimal classifier using only the features of the training nodes as input will have a positive error rate.  This can occur for example because nodes from different classes may have the same features, \emph{e.g.} papers in different subject areas in the Cora dataset may have the same bag-of-words as features, and thus cannot be distinguished by a purely feature-based classifier.  SAS likewise assumes the graph structure is not completely informative.  In particular, for a random edge from the graph, we assume there is probability $0 < \rho < 1$, which we call the \emph{homophily ratio}, that the endpoints of the edge belong to the same class.  A typical homophily ratio for a real-world dataset may be $\rho = 0.6$, providing useful but limited information for classification.  Given these assumptions, the basic idea for SAS is to first compute a probability vector for a node to belong to different classes, then leverage homophily to refine this estimate using the probability vectors of the node's neighbors.  The pseudocode for SAS is shown in Algorithm \ref{alg:sas} and described in detail below.  

\begin{algorithm}[t]
	\label{alg:sas}
	\SetKwInOut{Input}{Input}\SetKwInOut{Output}{Output}
	\caption{SAS Algorithm}
	\SetAlgoLined
	\Input{Graph $G$ \\
			Features of nodes  $\mathbf{X}_V$ \\
			Labels of training nodes $\mathbf{Y}_T$}
	Train $\mathbf{\Theta}$ in Eq. \ref{eq:mlp} on $(\mathbf{X}_T,\mathbf{Y}_T)$ using mini-batch SGD \\
    Apply trained MLP to $\mathbf{X}_V$ to obtain predictions $\mathbf{p}_1^{(0)}, \ldots, \mathbf{p}_N^{(0)}$ \\
    Apply Eq. \ref{eq-prop1} $K$ times to obtain final predictions $\mathbf{p}_1^{(K)}, \ldots, \mathbf{p}_N^{(K)}$ \\
	\Output{node $i$'s label $\gets$ max index in $\mathbf{p}_i^{(K)}$ }
\end{algorithm}

\subsection{Learning a Feature-Based Classifier}
In the first step of SAS, we ignore the graph information and train a node classifier using only feature information.  Specifically, we minimise the cross-entropy loss of a single or multi-layer perceptron (MLP) on the input features and labels of the training nodes.  
\begin{equation}
\label{eq:mlp}
\begin{aligned}
\mathbf{x}_i^{(0)}&=\mathbf{x}_i\\
\mathbf{x}_i^{(l)}&={\rm ReLU}(\mathbf{x}_i^{(l-1)}\mathbf{\Theta}^{(l)}),l\in\{1,\dots,L-1\}\\
\mathbf{p}_{i}&={\rm softmax}(\mathbf{x}_i^{(L-1)}\mathbf{\Theta}^{(L)})\\
{\rm minimize}\quad \mathcal{L}&=\frac{1}{|V_T|}\sum_{i\in T} -\mathbf{y}_i^\top\log (\mathbf{p}_{i})+\gamma \|\mathbf{\Theta}\|
\end{aligned}
\end{equation}
Here, $L$ denotes the number of layers in the MLP, $\mathbf{\Theta}^{(l)}$ are the parameters to be learned in layer $l$, $T$ is the set of training nodes, and $\gamma > 0$ is a regularization hyperparameter used to prevent overfitting.  We train this MLP using full or mini-batch gradient descent to minimize the loss.   

\subsection{Aggregation Algorithm}
After training the MLP in the previous step, we can apply it at all the nodes to produce a predicted label vector $\mathbf{p}_i$ for each node $i$.  The predicted label for $i$ is the index with the maximum value in $\mathbf{p}_i$.  However, these predictions may be incorrect, and we refine them using aggregation of predicted label vectors from the node's neighbors.  Specifically, each node $i$ performs $K > 0$ aggregations using the following:

\begin{equation}
\label{eq-prop1}
\mathbf{p}_i^{(k)}=\lambda_i  \mathbf{p}_i^{(0)}+(1-\lambda_i)\sum_{\{i,j\}\in \Tilde{E}} s_{ij} \mathbf{p}_i^{(k-1)}
\end{equation}
Here $\Tilde{E}$ is the set of edges with self-loops added, \emph{i.e.} $\Tilde{E} = E \cup \{(i,i) \, | \, i \in V\}$,  $\lambda_i \in [0,1]$ represents the amount of ``confidence" node $i$ has in its initial label prediction $\mathbf{p}_i^{(0)}$, and $s_{ij}$ is the edge weight between nodes $i$ and $j$, which we describe later.  Note that for labeled training nodes, we do not aggregate their one-hot labels $\mathbf{y}_i$ as in \cite{zhou2004learning,huang2020combining}, but instead aggregate their predicted labels $\mathbf{p}_i$.   The updates can also be expressed in matrix form as follows:
\begin{equation}
\mathbf{P}^{(k)}= \mathbf{\Lambda} \mathbf{P}^{(0)}+(\mathbf{I}-\mathbf{\Lambda})\Tilde{\mathbf{S}}\mathbf{P}^{(k-1)}
\label{alphaupdate}
\end{equation}
Here $\mathbf{\Lambda}={\rm diag}(\lambda_1,\dots,\lambda_N)$ and $\Tilde{\mathbf{S}}$ represents the \emph{coupling matrix} containing the weights $w_{ij}$.  SAS can use a number of different coupling matrices $\Tilde{\mathbf{S}}$, but we choose to use the weights from GCNs, defined as in 
Equation \ref{norm_adj} by $s_{ij}=1/\sqrt{(d_i+1)(d_j+1)}$, where $d_i$ is the degree of a node $i$, as a simple option. This matrix is widely used in many GNNs \cite{kipf2016semi,wu2019simplifying,chen2020simple} due its property as a low-pass filter in the graph spectral domain \cite{wu2019simplifying}.


For the matrix $\mathbf{\Lambda}$, we experimented with two options, either $\mathbf{\Lambda}=\mathbf{0}$ or  $\mathbf{\Lambda}=\alpha\mathbf{I}$, where $\mathbf{0}$ and $\mathbf{I}$ represent the zero and identify matrices respectively, and $\alpha \in (0,1]$ is a hyperparameter.  A positive $\alpha$ helps stabilize aggregation convergence, but requires a larger number of aggregations to obtain high accuracy.

Finally, note that after the last aggregation step, the entries in $\mathbf{p}_i^{(K)}$ do not necessarily sum to 1.  The final predicted label of node $i$ is the index with the maximum value in $\mathbf{p}_i^{(K)}$.

\subsection{Discussion}
SAS has several advantages compared to other algorithms.  The first is computational and memory efficiency.  Table \ref{complexity} compares the asymptotic complexity of SAS and several other algorithms for transductive end-to-end node classification, \emph{i.e.} classifying all the nodes in an input graph using a labeled training set.  SAS, SGC and GfNN have similar time complexity, while SAS's memory complexity is smaller by a significant constant factor on most datasets.  The three algorithms are also significantly faster and less memory intensive than GCNs and sampling based GNNs.  Sampling based algorithms in particular sometimes require a time consuming preprocessing step. 

Memory complexity is especially important in the context of graph learning, as memory intensive algorithms such as GCNs and GATs are often unable to run on even medium scale graphs with less than 1 million edges.  SAS's memory complexity during training is $O(LN_2 h)$, where $N_2$ is the size of a mini-batch during MLP training,  because it does not use any graph information.  In contrast, sampling based GNNs use more complex graph-based mini-batching techniques to achieve scalability, which reduces computational efficiency and also training stability.  SAS's memory complexity during the final inference step, when the trained MLP is applied at all nodes and the results are aggregated, is $O(|E|C)$ because nodes only propagate label vectors with size $O(C)$.  In contrast, SGC, GfNN and GNNs propagate feature vectors and other latent representations, which usually have significantly higher dimensionality.  An example of this effect is discussed in Section \ref{real-benchmarks}.

\begin{table}[t]
	\caption{Asymptotic time and memory complexity for training and validation. $n$ is the number of training epochs, $h$ is the dimension of the hidden layer, $L$ is the number of layers in the MLP classifier, $K$ is the number of times we aggregate, $N$ and $M$ are the number of nodes and edges in the graph, resp. For sampling methods, $b$ is the neighbor sample size, $N_1$ and $M_1$ are the number of nodes and edges in the sampled subgraph, resp.  $N_2$ is the size of a mini-batch during MLP training.}\label{complexity}
	\begin{center}
		\begin{tabular}{|l|l|l|}
			\hline
			&  Training and validation time  &  Memory (per batch)\\
			\hline
			GCN &    $O(nKMh+nKNh^2)$&$O(KMh)$ \\
			GraphSAGE &  $O(nb^K Nh^2)$ & $O(b^K N_1 h)$\\
			GraphSAINT &  $O(nKMh+nKNh^2)$&$O(KM_1 h)$\\
			SGC/GfNN & $O(KMD)+O(nLNh^2)$ &$O(LN_2 h)$\\
			\hline
			SAS & $O(KMC)+O(nLNh^2)$ &$O(LN_2 h)$\\
			\hline
		\end{tabular}
	\end{center}
\end{table}

As we show in Section \ref{sec-exp}, SAS's accuracy is competitive with computationally more expensive GNNs and higher than other two-stage algorithms with similar computational cost.  We discuss SAS's improved accuracy further in Section \ref{sec:sythetic}.

\section{Experiments and Discussions}
\label{sec-exp}
In this section, we evaluate SAS for its accuracy and computational performance on a number of real-world datasets.  We also present synthetic datasets to explain the reason why SAS achieves higher accuracy than current two-phase algorithms.  
\begin{table}[!t]
\caption{Benchmark dataset statistics.  $D$ is the dimension of input features, $C$ is the number of classes.}\label{statistics}
\begin{center}
\begin{tabular}{|l|l l l l|l|}
\hline
Dataset &  $N$ & $|E|$ & $D$ & $C$ & Train / Val / Test\\
\hline
Cora &  2,708  & 5,429 & 1,433 & 7 & 5\% / 18\% / 37\% \\
Citeseer &  3,327 & 4,732 & 3,703 & 6 & 4\% / 15\% / 30\% \\
PubMed & 19,717 & 44,338 & 500 & 3 & 0.3\% / 2.5\% / 5\% \\
Flickr & 89,250 & 899,756 & 500 & 7 & 50\% / 25\% / 25\% \\
ogbn-arxiv & 169,343 & 1,166,243 & 128 & 40 & 54\% / 17\% / 29\% \\
ogbn-products & 2,449,029 & 61,859,140 & 100 & 47 & 10\% / 2\% / 88\% \\
Reddit & 232,965 & 114,848,857 & 602 & 41 & 66\% / 10\% / 24\% \\
\hline
\end{tabular}
\end{center}
\end{table}

\subsection{Performance on Benchmark Datasets}
\label{real-benchmarks}

\subsubsection{Datasets}
The datasets we evaluate are shown in Table \ref{statistics}.  These include several small citation networks Cora, Citeseer and PubMed.  The training / validation / test splits of these datasets are provided by \cite{yang2016revisiting}.  We also consider two medium scale datasets, the Flickr dataset from \cite{zeng2019graphsaint} and another citation network ogbn-arxiv from \cite{hu2020open}.  Lastly, we test two large datasets,  the original Reddit dataset from \cite{hamilton2017inductive}\footnote{Note that the version of Reddit we use is denser than those used in \cite{wu2019simplifying,zeng2019graphsaint}}, and the ogbn-products dataset is from \cite{hu2020open}.  Flickr and Reddit are inductive tasks, while all the others are transductive.

\begin{table}[t]
\caption{Accuracy of SAS compared to other methods. The highest accuracy on each dataset is marked in bold.  }\label{table:performace}
\begin{center}
\begin{threeparttable}
 \begin{tabular}{|l|ccccccc|}
\hline
Method &  Cora & Citeseer & PubMed & Arxiv & Products & Flickr & Reddit\\
\hline
GCN &  0.815  & 0.703 & 0.790 & 0.717 & OOM$^\dagger$ & 0.492 & OOM$^\dagger$ \\
GAT &  0.830 & $\mathbf{0.725}$  & 0.790 & $\mathbf{0.735}$ & OOM$^\dagger$ & $0.423^*$ & OOM$^\dagger$ \\
APPNP & $\mathbf{0.833}$ & 0.718 & 0.801 & $0.667^*$ & OOM$^\dagger$ & $0.425^*$ & OOM$^\dagger$\\
ClusterGCN & - & - & - & ERR$^\ddagger$ & 0.752 & 0.481 & 0.954 \\
GraphSAGE & - & - & - & $0.659^*$ & $\mathbf{0.783}$ & 0.501 & 0.954 \\
GraphSAINT & - & - & - & $0.713^*$ & 0.773 & 0.511 & $\mathbf{0.966}$ \\
SGC & 0.810 & 0.719 & 0.789 & $0.619^*$ & $0.653^*$ & $0.507^*$ & $0.937^*$ \\
GfNN & $0.815^*$ & $0.710^*$ & $0.794^*$ & $0.702^*$ & $0.720^*$ & $0.505^*$ & $0.940^*$\\

\hline
SAS-A & 0.820 & 0.721 & 0.813 & 0.712 & 0.759 & $\mathbf{0.521}$ & 0.944\\
SAS-B & 0.832 & $\mathbf{0.725}$ & $\mathbf{0.814}$ & 0.713 & 0.755 & 0.505 & 0.951\\
\hline
\end{tabular}
 \begin{tablenotes}
       \footnotesize
        \item[*] Result from our experiment.
       \item[$\dagger$] GPU out-of-memory error.
        \item[$\ddagger$] Runtime error.
     \end{tablenotes}
\end{threeparttable}

\end{center}
\end{table}

\subsubsection{Setup} 
We test two versions of SAS which differ in the use of a residual connection.  The first version SAS-A sets $\mathbf{\Lambda}=\mathbf{0}$ in Equation \ref{alphaupdate}, while SAS-B sets $\mathbf{\Lambda}=\alpha \mathbf{I}$ for a hyperparameter $\alpha$. Model parameters are optimized by minimizing the cross-entropy loss using full or mini-batch gradient descent with the Adam optimizer \cite{kingma2014adam}. In order to reduce overfitting, we apply the standard regularization techniques of weight decay, dropout and batch normalization \cite{ioffe2015batch}.  Hyperparameters were optimized automatically using Bayesian optimization provided in the AX Platform\footnote{https://ax.dev/}.

\subsubsection{Baselines} We compare our model with a number of state-of-the-art GNNs, including GCN \cite{kipf2016semi}, GAT \cite{velivckovic2017graph}, ClusterGCN \cite{chiang2019cluster}, GraphSAGE \cite{hamilton2017inductive}, and GraphSAINT \cite{zeng2019graphsaint}, APPNP \cite{klicpera2018predict}, SGC \cite{wu2019simplifying} and GfNN \cite{nt2019revisiting}.  GNN, GAT and APPNP are not optimized for scalability, and encounter OOM (out-of-memory) errors on the ogbn-products and Reddit datasets.  We do not report results for ClusterGCN, GraphSAGE and GraphSAINT on the small citation networks, as these methods are designed for large networks.  Also, ClusterGCN crashed on the ogbn-arxiv dataset.  We used original implementations provided by the authors where possible.  


\subsubsection{Implementation Details} All experiments were done on Ubuntu 18.04 on an Intel(R) Xeon(R) Silver 4114 @ 2.20GHz CPU with 128 GB of RAM and an Nvidia GTX 1080 Ti GPU with 11GB of RAM.  We implemented SAS in Pytorch and PyG \cite{fey2019fast}.

\subsubsection{Accuracy}
Table \ref{table:performace} shows the classification accuracy of the models.  We report micro-f1 score for the inductive datasets Flickr and Reddit, as in \cite{hamilton2017inductive}.  Some results in Table \ref{table:performace} are taken from the original papers or followup works, while results marked with $\star$ are based on our own experiments.  In the latter case, to ensure fairness, we optimized model hyperparameters using the AX Platform.  

We highlight some of the main findings.  First, for the small citation datasets Cora, Citeseer and PubMed, we achieve equal or sometimes higher accuracy than  state-of-the-art GNNs, suggesting that some of the complex operations they perform may not be necessary.  Second, as one of the key goals for our design is computational efficiency, we focus on comparing SAS with two other methods optimized for efficiency, namely SGC and GfNN.  SAS exceeds the accuracy of these methods on all the datasets except Flicker.  As the main difference between SAS and these methods is the order of the transformation and aggregation steps, this indicates that doing transformation first improves accuracy.  Third, while SAS-B was generally more accurate than SAS-A, it had lower accuracy on Flickr.  This may be because the MLP in the first stage of SAS achieved quite low accuracy on Flickr, around 40\%.  In this case, it may be better for nodes not to rely too much on their own MLP based predictions, but rather put more weight on the aggregated predictions of their neighbors.  SAS-A does this by avoiding residual connections, possibly leading to its higher accuracy.  

Lastly, compared to the state-of-the-art graph sampling based methods GraphSAGE and GraphSAINT, SAS shows comparable but lower accuracy.  However, as we show later, these methods incur substantially higher running times.  In addition, we currently train SAS using full batch training, as we found mini-batch training was unnecessary for even the largest graphs.  However, GraphSAGE and GraphSAINT are trained using minibatches, and this may possibly improve their accuracy \cite{hamilton2017inductive,zeng2019graphsaint} by simulating an edge-dropout effect.  We leave evaluation of mini-batch training for SAS as future work.

\subsubsection{Runtime and Memory Usage}
The running time for SAS and several other methods are shown in Table \ref{table:time}. SAS is significantly faster than GraphSAGE and GraphSAINT, is slightly faster than GfNN, and is substantially slower than SGC.  The reason for the latter is that SGC uses only a single linear layer in its transformation step, resulting in very fast training.  The drawback to this simplicity is that SGC achieves significantly lower accuracy. SAS and GfNN both use 3 layer MLPs with a size 256 hidden dimension, which takes over $20 \times$ longer to train than SGC.  Table \ref{table:time} breaks down the overall running time for each method into several components, including preprocessing, training, inference and postprocessing.  Due to lack of space, we do not describe these components in detail, but note that preprocessing, training and inference take up the bulk of the running time in GraphSAGE and GraphSAINT, while SAS's running time is dominated by the time to train its MLP.  In addition, GraphSAINT first performs an expensive normalization step to compute sampling probabilities, which we did not count, but which would more than triple its running time if included.

\begin{figure}[!t]
\centering
\includegraphics[width = 1\textwidth,clip]{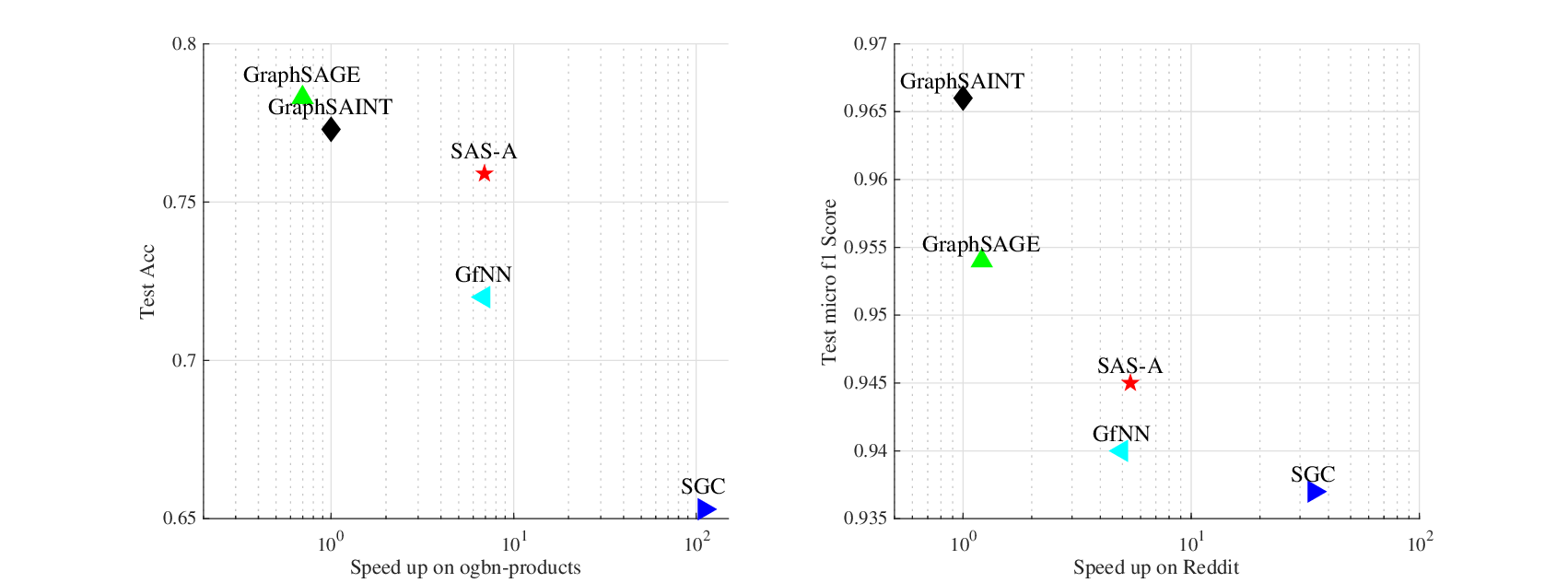}
\caption{Accuracy compared to speedup (with GraphSAINT's running time as a baseline) on the ogbn-products (left) and Reddit (right) datasets.}
\label{figure:tradeoff}
\end{figure}

Since SAS and GfNN both use 3-layer MLPs for transformation, they have similar training times.  However, SAS is significantly faster in the aggregation stage, as shown in Table \ref{table:time}.  Here, SAS applies the trained MLP at all nodes to produce label prediction vectors, then propagates these vectors to produce the final predicted label.  For GfNN, aggregation is done before MLP training, and involves propagating the nodes' initial feature vectors.  However, feature vectors are typically much larger than label vectors; for example, Reddit feature vectors have dimension 602, while the label vector has dimension 41.  As a result, GfNN sends much more data during the aggregation stage, which results in longer execution time and more memory usage. At the same time, since the overall performance of GraphSAINT is best in Table \ref{table:performace}, we refer it as the baseline and report the performance and relative time comparison for all the other methods compared with GraphSAINT over ogbn-products and Reddit compared with GraphSAINT in Figure \ref{figure:tradeoff}. 

\begin{table}[!ht]
    \caption{Wallclock time for aggregation / preprocessing , training, inference, and total time, on the ogbn-products and Reddit datasets.  Results are averaged over 5 runs. }\label{table:time}
   \centering
   \begin{threeparttable}
    \centering
    \begin{tabular}{|ll|c|c|c|c|}
    \hline
    Dataset                           & Method     & Aggregation / preprocessing & Training & Inference & Total        \\ \hline
    \multicolumn{1}{|l}{}              & SGC        & 1.706            & 5.313    & 0.0401    & 7.059   \\
                                      & GfNN       & 1.714            & 111.4    & 0.6263    & 113.7  \\
    \multicolumn{1}{|l}{ogbn-products} & GraphSAGE  & 82.12            & 925.5    & 110.7     & 1118  \\
                                      & GraphSAINT & 59.20            & 605.7    & 115.9     & 780.8     \\
                                      & SAS-A    & 1.694            & 110.5    & 0.5312    & 112.7 \\ \hline \hline
                                      & SGC        & 6.946            & 6.039    & 0.043     & 13.03 \\
                                      & GfNN       & 6.721            & 83.64    & 0.396     & 90.76 \\
    \multicolumn{1}{|l}{Reddit}        & GraphSAGE  & 35.59            & 303.5    & 33.52     & 372.6 \\
                                      & GraphSAINT & 17.18            & 380.7    & 52.19     & 450.1    \\
                                      & SAS-A    & 1.708            & 81.02    & 0.394     & 83.12 \\ \hline
    \end{tabular}
     
   \end{threeparttable}
  \end{table}


\subsubsection{Number of Aggregations}  
\begin{figure}[t]
	\begin{minipage}[t]{0.45\textwidth}
		\centering
		\includegraphics[scale=0.5]{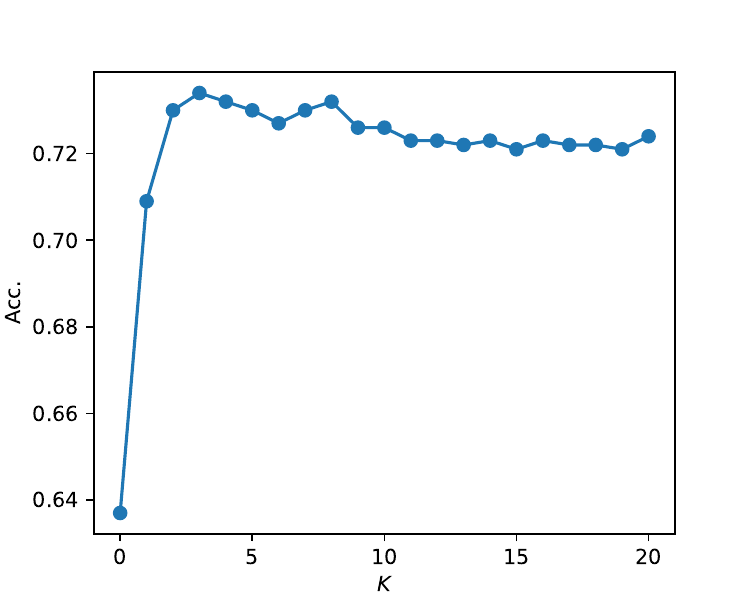}
	\end{minipage}
	\qquad
	\begin{minipage}[t]{0.45\textwidth}
		\centering
		\includegraphics[scale=0.5]{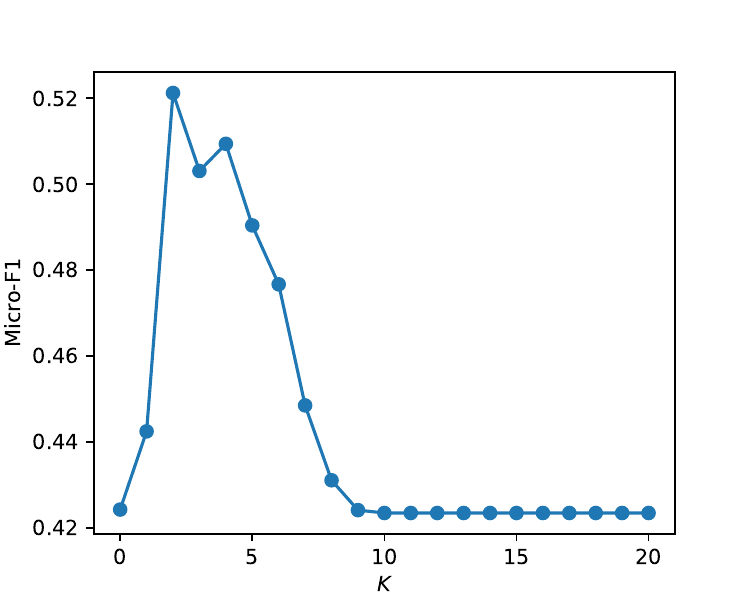}
	\end{minipage}
	\caption{Accuracy or micro-f1 score as a function of the number of aggregations $K$ on Citeseer (left) and Flickr (right). $K=0$ represents the result of the base classifier.
		\label{fig:acc}}
\end{figure}

We next examine how the number of aggregations $K$ affects SAS-A's accuracy.  Figure \ref{fig:acc} shows the accuracy as we vary $K$ for two datasets Citeseer and Flickr.  Citeseer maintains largely consistent accuracy even for large $K$, whereas Flickr's accuracy peaks at $K=2$ and quickly drops down to the accuracy of the MLP classifier from the transformation step for large $K$.  One reason for this is \emph{over-smoothing}, \emph{i.e.} after a large number of aggregations each node's predicted label vector will converge to a vector only related to the node's degree \cite{li2018deeper}.  In Flickr this convergence occurs more quickly because the average degree in the graph is around 10, whereas in Citeseer the average degree is around 1.5.

Choosing the right number of aggregations has a significant effect on accuracy.  In SAS, we can find a good value for $K$ by looking at the accuracy on the validation set after performing each aggregation, and stopping after the accuracy starts to decrease.  This incurs low computational cost, since each validation and aggregation step is fast.  On the other hand, SGC and GfNN cannot efficiently perform this tuning on $K$.  This is because they first perform $K$ aggregations, then train an MLP based on the aggregated features.  Thus, these methods need to retrain the MLP for each different value of $K$ they test, which is inefficient because MLP training takes far longer than validation and aggregation.

\begin{figure}[!h]
    \centering
    \includegraphics[width = 0.9\textwidth,trim=125 25 100 25,clip]{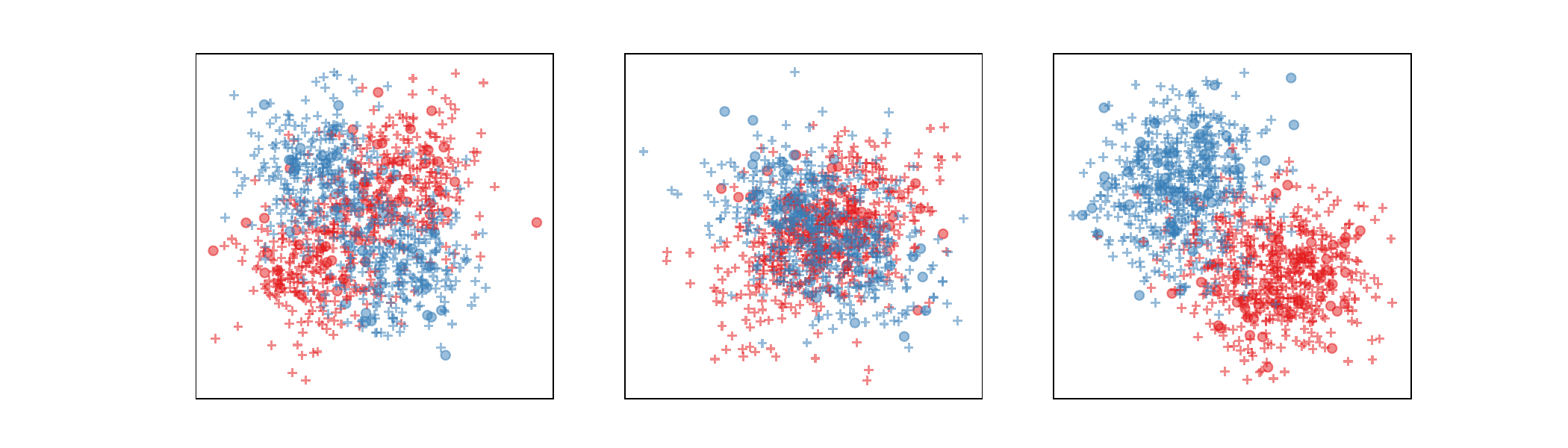}
    \caption{Synthetic XOR dataset.  Classes $A$ and $B$ are marked in red and blue respectively.  Training nodes ($|T| = 100$) are marked with '$\bullet$', test nodes ($|V \backslash T|$ = 1000) are marked with '$+$'.  (Left)  Distribution of node features.  (Middle) Features after one aggregation.  (Right) Latent representation after MLP then aggregation; note that the MLP outputs a 2D vector for each node.}
    \label{fig:xor}
\end{figure}

\begin{figure}[!h]
    \centering
    \includegraphics[width = 0.9\textwidth,trim=125 25 100 25,clip]{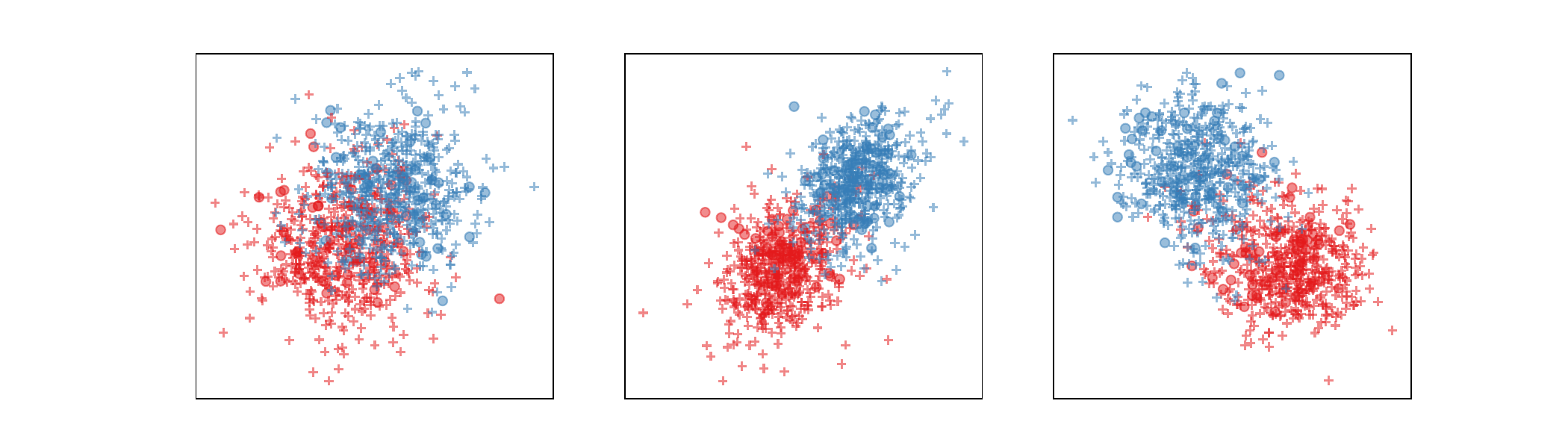}
    \caption{Synthetic Gaussian dataset, using the same settings as in Figure \ref{fig:xor}. (Left)  Distribution of node features.  (Middle) Features after one aggregation.  (Right) Latent representation after MLP then aggregation; note that the MLP outputs a 2D vector for each node.}
    \label{fig:gaus}
\end{figure}

\subsection{Synthetic Datasets}
\label{sec:sythetic}
Finally, we consider two synthetic datasets which highlight the reasons for SAS's improved accuracy compared to other two-stage algorithms.  

\subsubsection{Experimental Setup} The first dataset captures a 2D noisy exclusive OR (XOR) function.  It consists of nodes with features $(x_0, x_1)$ and belonging to a class $A$ or $B$.  Class $A$ contains points generated from two 2D Gaussian distributions with means at $(1,1)$ and $(-1, -1)$ and covariance matrix $\mathbf{\Sigma}={\rm diag}(0.75,0.75)$, while class $B$ consists of two 2D Gaussian distributions with means at $(1, -1)$ and $(-1, 1)$ and covariance matrix $\mathbf{\Sigma}={\rm diag}(0.75,0.75)$.  An example of the dataset is shown in Figure \ref{fig:xor}.   

The second dataset also contains two classes $A$ and $B$, but now both classes are 2D Gaussian distributions centered at $(1,1)$ and $(-1, -1)$ respectively and with the same covariance matrix $\mathbf{\Sigma}'={\rm diag}(3,3)$.  An example of the dataset is shown in Figure \ref{fig:gaus}.   


Classification accuracy can be improved if we use graph information in addition to the features.  To this end, we created a graph on the points in the dataset with 80\% homophily ratio, which is roughly equal to the homophily ratio on the small citation networks in Section \ref{real-benchmarks}.  To do this, we first formed a Erdos–Rényi random graph with average degree 3 on the nodes in class $A$ and separately on the nodes in class $B$, resulting in a total of $e$ edges.  We then formed a random bipartite graph using the method in between the nodes in classes $A$ and $B$ with a total of $\frac{e}{4}$ edges.

\subsubsection{Baselines} We compared SAS's accuracy with a number of other algorithms in which we vary the number of transformation and aggregation steps and their ordering.  We use $T$ and $A$ to denote the transformation and aggregration steps and use these to represent each algorithm.  For example, T-T-A-A represents an algorithm which first performs two transformation steps then two aggregation steps, and is equivalent to SAS using a 2-layer MLP and 2-hop aggregation.  Similarly, A-A-T represents an algorithm which first performs two aggregation steps followed by a transformation, and is equivalent to SGC.  The other baselines are shown in Table \ref{xor_acc}.


\subsubsection{Analysis} 
We now analyze the results in Table \ref{xor_acc}, starting with the Gaussian dataset.  Here, the optimal classification accuracy based only on features is 80.6\%, and a two layer MLP (model T-T) nearly achieves this.  Also, all models which perform at least one step of aggregation exceed the optimal feature-based accuracy, with more aggregation steps generally increasing accuracy, indicating the utility of graph structure for classification.  

Next, we consider the XOR dataset.  Here, the optimal feature-based accuracy is 79.7\%, and a two-layer MLP achieves 70.8\% accuracy.  In addition, by performing one or two aggregation steps after the MLP transformation, the accuracy can be substantially improved, to 87\%.  However, performing aggregation before the transformation step results in significantly lower accuracy.  In particular, the A-A-T model, corresponding to the SGC algorithm, performs only slightly better than chance.  A-A-T-T, corresponding to GfNN with two aggregation steps, has 55.5\% accuracy, while A-T-T's accuracy is slightly better.  

To understand the reason for these results, consider the left and middle panels in Figure \ref{fig:xor}, showing the original XOR distribution, and the distribution after one aggregation step.  We see that directly aggregating on the features of the XOR dataset results in heavy mixing of the two classes and increases the number of samples which cross (\emph{i.e.} lie on the wrong side of) the optimal decision boundary given by ${\rm sign}(x_0 \cdot x_1)$, causing a decrease in the maximum achievable accuracy.  Additional propagation steps will result in even greater mixing and further decreases accuracy.  Meanwhile, by transforming the data first using an MLP, as in the third panel in Figure \ref{fig:xor}, we map the points so that only about 30\% of the transformed points cross the MLP decision boundary.  Subsequent aggregation steps further improve the accuracy, because crossing points likely have several non-crossing neighbors due to the 80\% homophily ratio, and aggregating label predictions from these neighbors helps correct the label predictions of the crossing nodes.    

Lastly, we compare the A-A-T and A-A-T-T models, corresponding to SGC and GfNN respectively.  On the Gaussian dataset, whose optimal decision boundary is a straight line, these models achieve nearly the same accuracy, indicating that a single layer perceptron is sufficient for this simple decision boundary.  However, on the XOR dataset, whose optimal decision boundary is nonlinear, A-A-T-T achieves substantially higher accuracy than A-A-T, indicating that deeper MLPs are needed for nonlinear decision boundaries.  This justifies the use of multi-layer MLPs in SAS, especially on the complex datasets described in Section \ref{real-benchmarks}.  

\begin{table}[t]
\caption{Accuracy for different interleavings of permutations of aggregation and transformation steps on the XOR and Gaussian datasets.  \emph{Random} and \emph{Optimal} the accuracy of random and optimal classifier respectively.  }\label{xor_acc}
\begin{center}
\begin{tabular}{|l|l|l|}
\hline
Pipeline    & XOR  & Gaussian \\ \hline
Random      & 0.5   & 0.5\\
Optimal & 0.797 & 0.806\\ \hline
A-A-T (SGC) & 0.518 & 0.927\\
A-A-T-T (GfNN)    & 0.555 &0.928\\
A-T-T (GfNN) & 0.562 &0.884\\
T-T (MLP)   & 0.708 &0.804\\ \hline
T-T-A (SAS)      & 0.807 &0.895\\
T-T-A-A  (SAS)  & 0.870 &0.931\\ \hline
\end{tabular}
\end{center}
\end{table}

\section{Conclusion}
\label{sec-concl}
In this paper, we proposed SAS, a simple, accurate and scalable node classification algorithm.  SAS first trains an MLP for label prediction using node features, then performs multihop aggregation of predicted labels to improve accuracy.  Our model has comparable accuracy as state-of-the-art GNNs on a variety of real world datasets, while achieving significantly higher performance and scaling to graphs with over one hundred million edges.  We also analyzed the reason SAS achieves higher accuracy compared to other two-stage algorithms by showing the importance of transforming features before aggregating them.  In the future, we will study how the number of aggregations affects prediction accuracy, and also try to relate graph structure to the classifier's architecture.  
\bibliographystyle{splncs04}
\bibliography{mybibliography}
%






\end{document}